\documentclass[letterpaper, 10 pt, journal, twoside]{IEEEtran}
\IEEEoverridecommandlockouts                              

\usepackage{graphics} 
\usepackage{amsmath} 
\usepackage{amssymb}  
\usepackage{siunitx}
\usepackage{mathtools}
\usepackage{booktabs}
\usepackage[ruled]{algorithm2e}
\usepackage{subcaption} 
\usepackage{adjustbox}
\usepackage{pbox}
\usepackage{dblfloatfix}
\usepackage{graphicx}
\usepackage{xcolor}

\begin{document}

\title{
An Efficient Sampling-based Method for Online Informative Path Planning in Unknown Environments
}


\author{Lukas Schmid$^1$, Michael Pantic$^1$, Raghav Khanna$^1$, Lionel Ott$^2$, Roland Siegwart$^1$, and Juan Nieto$^1$

\thanks{$^1$ The authors are with Autonomous Systems Lab, Department of Mechanical and Process Engineering, ETH Z\"urich, 8092 Z\"urich, Switzerland
{\tt\small schmluk@mavt.ethz.ch}}%
\thanks{$^2$ The author is with The University of Sydney, NSW 2006, Australia}%
}%

\maketitle

\begin{abstract}

The ability to plan informative paths online is essential to robot autonomy. In particular, sampling-based approaches are often used as they are capable of using arbitrary information gain formulations. However, they are prone to local minima, resulting in sub-optimal trajectories, and sometimes do not reach global coverage. In this paper, we present a new RRT*-inspired online informative path planning algorithm. Our method continuously expands a single tree of candidate trajectories and rewires nodes to maintain the tree and refine intermediate paths. This allows the algorithm to achieve global coverage and maximize the utility of a path in a global context, using a single objective function. We demonstrate the algorithm's capabilities in the applications of autonomous indoor exploration as well as accurate Truncated Signed Distance Field (TSDF)-based 3D reconstruction on-board a Micro Aerial Vehicle (MAV). We study the impact of commonly used information gain and cost formulations in these scenarios and propose a novel TSDF-based 3D reconstruction gain and cost-utility formulation. Detailed evaluation in realistic simulation environments show that our approach outperforms sampling-based state of the art methods in these tasks.
Experiments on a real MAV demonstrate the ability of our method to robustly plan in real-time, exploring an indoor environment with on-board sensing and computation. We make our framework available for future research.

\end{abstract}

\section{INTRODUCTION}
\IEEEPARstart{I}{n} recent years, mobile robots and in particular, MAVs have shown increasingly high levels of autonomy and can be employed in an ever-growing number of applications. A crucial component to leveraging their full potential is the ability to autonomously plan and execute informative paths in a priori unknown environments. 

In particular, sampling-based methods are widely used, as various information gains can be directly computed from the map without imposing additional constraints on the choice of gain formulation.
These methods have proven successful in a variety of scenarios, including volumetric exploration \cite{nbvp_bircher}, surface inspection \cite{Bircher2018}, object search \cite{nbvp_object_search}, weed classification \cite{popovic2017ipp}, infrastructure modeling \cite{Yoder2016}, and 3D reconstruction \cite{song_3d_rec}.

In the informative path planning (IPP) problem, a robot is required to generate a path that maximizes the information gathered, i.e. maximizes an objective function, about its environment.
To solve the IPP problem when the environment is unknown, robots are required to identify promising paths online based on the limited information available at each time step. 
This requires paths to be continuously adapted to the current map. 
Furthermore, the robot needs to reason over a potentially large map space to escape local minima and strive for globally optimal plans.
In many cases, both tasks need to be performed with the limited computational resources available on-board the robot. 
Since the operational time of mobile robots is typically limited by the battery life, efficient computation of informative trajectories is of major importance.

\begin{figure}[t]
     \centering
     \includegraphics[height=5.7cm]{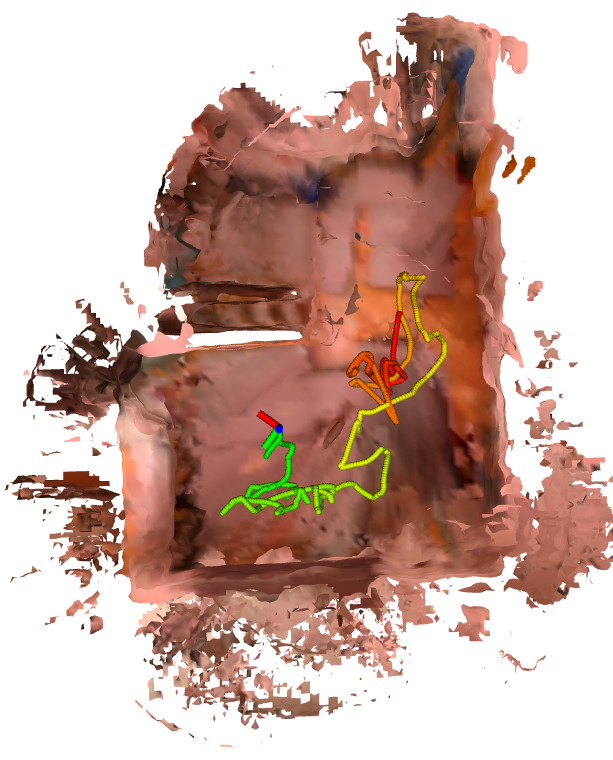}
     \hfill
     \includegraphics[height=5.7cm]{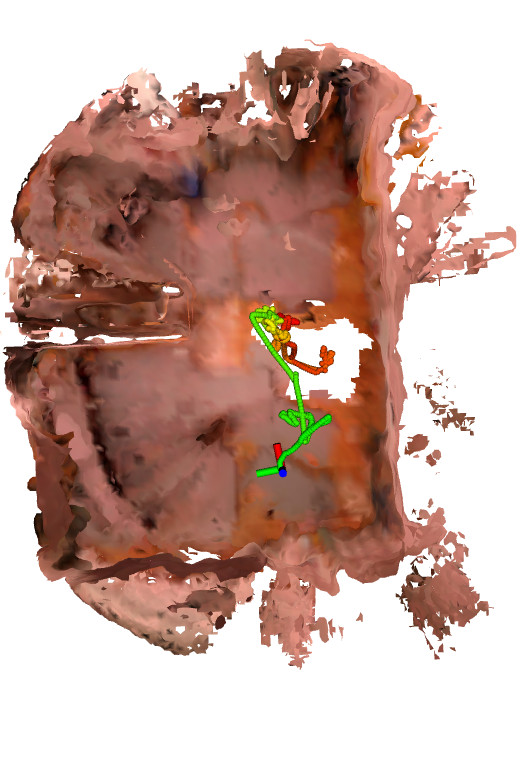}

    \caption{Qualitative comparison of our method (left) and RH-NBVP \cite{nbvp_bircher} (right) on a real MAV. The trajectories are colored from red at $t=0$ \si{\minute} to green at $t=3$ \si{\minute}. Our method explores the full room and passes the obstacle in the bottom center, while RH-NBVP leaves a hole in the floor. Our gain focuses on areas of high expected error, leading the MAV to revisit surfaces and resulting in better resolved corners and flatter walls.}
    \label{fig:real_room}
\end{figure}

Traditional sampling-based online IPP approaches, based on repeatedly expanding a rapidly-exploring random tree (RRT) \cite{nbvp_bircher, nbvp_object_search, history_nbvp, yang2013gaussian, nbvp_uncertainty}, iteratively sample feasible paths from the current robot pose, storing them in a tree structure, and execute the beginning of the best branch.
However, these approaches have two important disadvantages. 
First, large parts of the tree are discarded every iteration, as either only the best branch is executed or the tree is expanded from scratch.
This can be computationally expensive; for example, computing volumetric information gains can account for up to 95\% of a planner's run-time.
Second, due to the limited horizon look-ahead, they tend to select locally optimal solutions, resulting in sub-optimal paths or even getting stuck in a dead-end. 

To address these challenges, we adapt ideas from optimal motion planning to the online informative path planning problem.  
Similar to the RRT* \cite{rrt_star} and real-time-RRT* \cite{rt_rrt_star} algorithms, we propose the rewiring of nodes according to their utility. 
In this way, non-executed nodes and their sub-trees are kept alive.
In combination with an adaptive updating policy, an ever-growing tree can be expanded and maintained while traversing it. 
This allows the algorithm to use a single objective function instead of switching operation modes, maintaining a large tree for global coverage, and simultaneously refining the planned trajectory.

We evaluate the proposed approach in two representative applications, (i) exploration of indoor environments, and (ii) accurate TSDF-based 3D reconstruction of buildings. 
Because the algorithm is strongly governed by a single objective function, we study the impact of commonly used information gains and costs in these scenarios. To address the problems identified from this, we present a gain formulation selecting viewpoints for accurate TSDF-based 3D reconstruction. Additionally, we introduce a parameter-free method to combine costs and gains into a total utility value.
Quantitative evaluation is carried out in complex and photo-realistic simulation environments, while experiments on a flying robot validate the method in the physical world. This work makes the following contributions to the state of the art:
\begin{itemize}
    \item We present a new RRT*-inspired online IPP algorithm, that continuously expands and maintains a single tree, reaching global coverage and refining trajectories to maximize their utility.
    \item We propose an information gain for accurate TSDF-based 3D reconstruction under uncertainty, as well as an efficiency-based utility formulation that balances gain and cost without additional tuning. 
    \item The algorithm is open-sourced with a modular framework for sampling-based online IPP design\footnote{\label{foot:mav_active}https://github.com/ethz-asl/mav\_active\_3d\_planning}. We also make the simulation environment and scenarios available for benchmarking and future research\footnote{\label{foot:uecvros}https://github.com/ethz-asl/unreal\_cv\_ros}.
\end{itemize}


\section{RELATED WORK}
Online planning of informative paths in unknown environments has attracted a lot of interest, primarily in the field of exploration planning. 
Here, the objective is to map unknown space in a region of interest completely and quickly. 
Although there exists a rich variety of approaches \cite{aut_expl_comparison}, the majority can be split into frontier-based and sampling-based methods. 
Frontier-based methods \cite{yamauchi1997frontier} identify the boundaries between known and unknown space in the map and repeatedly choose one such frontier as the goal, eventually resulting in complete exploration. 
This approach has, among others, been extended to high speed flight for fast exploration \cite{rapid}, but is difficult to adapt to other tasks.
Sampling-based methods typically sample view configurations in a Next-Best-View (NBV) fashion \cite{nbvp_bircher}.
These have the large advantage of allowing any kind of gain formulation \cite{nbvp_uncertainty,nbvp_object_search}. 
For this reason, we focus on sampling-based methods in this work. 

Recent approaches leverage the capabilities of sampling-based methods and employ additional planning stages to escape local minima or address sub-optimal trajectories.
Corah et al. \cite{gmm_multirobot_expl} generate a tree of motion primitives in a fixed horizon fashion to identify promising local trajectories.
Areas of potential information gain in the map are approximated by a global library of uniformly sampled views.
Local minima are then escaped by penalizing local trajectories that end far from library views of a fixed minimum gain. However, the computed gains are not used to find better global targets.
Another approach \cite{charrow_ITplanning} utilizes a frontier method to detect global goals and supplements these with motion primitives for local exploration. 
The most promising trajectory is then refined in a gradient-based manner to maximize their information objective.
Meng et al. \cite{meng_2stage_expl} present a two stage approach that samples viewpoints near frontiers as candidates for coverage. 
To account for the global quality of the exploration path, viewpoints are cast into a fixed-start open traveling salesman problem to obtain a globally optimal traversal path. However, to remain tractable the available viewpoints are pruned and the trajectory lengths are approximated using a heuristic gain formulation.

A different family of approaches focuses on expanding an RRT of viewpoints, as introduced by Bircher et al. \cite{nbvp_bircher}. 
In a receding horizon (RH) fashion, a tree of views is sampled, and the first node of the best branch is executed.
To escape local minima more efficiently, Witting et al.~\cite{history_nbvp} keep track of the planner's history as potential areas for reseeding the tree.
Similarly, Selin et al. \cite{Selin_nbv_fron} use the RH-NBV planner for local exploration and a frontier-based method for global goal selection.

While volumetric exploration algorithms are able to fully discover and reconstruct large outdoor structures \cite{nbvp_bircher, meng_2stage_expl, Selin_nbv_fron}, more specialized surface-based methods exist.
Yorder et al. \cite{Yoder2016} introduce the concept of surface frontiers. Assuming a single connected object of interest, detecting unknown voxels next to surfaces results in full coverage. 
In a NBV fashion, informative paths are planned, where the sampling space for viewpoints is significantly reduced by uniformly sampling on an offset transform of the currently available model.
An approach that accounts for the surface quality of the model is presented in \cite{song_3d_rec}. 
While RRT-based volumetric exploration is used to identify goal poses, intermediate paths are refined to inspect surfaces and explore surroundings. 
As a measure of quality, the confidence of a surface is defined as the average of the TSDF weights of neighboring points.
For points below a threshold, viewpoints that can observe the low confidence point are sampled outwards and added to the set of candidates.
However, due to the constant TSDF update weight used, high weights do not necessarily translate into high quality and no metric accuracy analysis is provided.

The approaches described above can escape local minima by utilizing a second planning stage.
However, this requires them to switch operation modes, and the notion of frontiers may not be meaningful for arbitrary information gain formulations.
In contrast to this, we propose the expansion of a single large tree, allowing it to work with a single objective function and to consider the utility of both local and global paths simultaneously.


\section{PROPOSED ALGORITHM}
\begin{figure}
     \centering
     \includegraphics[width=\linewidth]{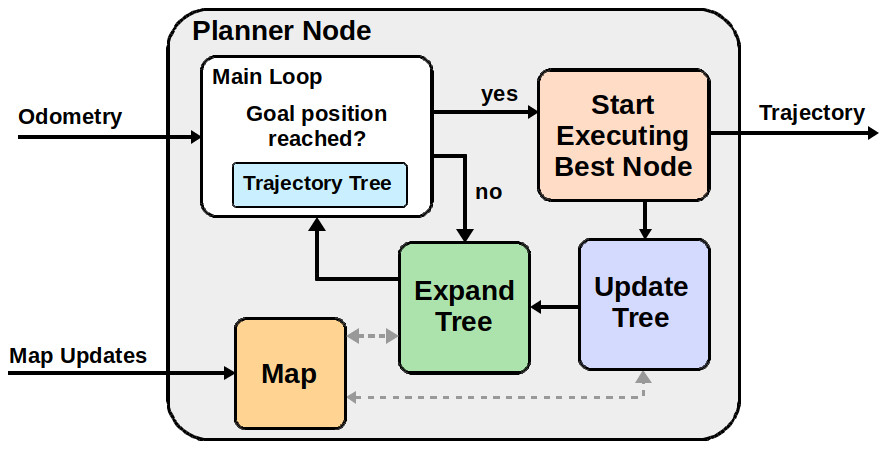}
     \caption{Planner overview. The tree is expanded until the current trajectory has finished execution. The tree is then updated to the new map and non-executed nodes are rewired to keep them alive.}
     \label{fig:planner_overview}
\end{figure}

The central idea of our approach is to continuously expand, maintain, and improve a single trajectory tree while simultaneously executing it.
An overview of the algorithm is given in Fig.~\ref{fig:planner_overview}.
In a receding horizon fashion, the tree is expanded until the current node has been completed. The trajectory of the best adjacent node is then requested and the tree updated.
Because the tree, i.e. the look-ahead, is generally kept alive, the next node is requested immediately, resulting in an adaptive horizon length.

\subsection{Tree Structure}
We represent the tree $\mathcal{G}=(\mathcal{V},\mathcal{E})$ as a set of vertices $\mathcal{V} = \{V_i\}_{i=1,\dots,n}$ and connectivity information $\mathcal{E}$, with the root located at the robot's current position. Each node $V_i$ consists of an associated trajectory $\tau$, gain $g$, cost $c$, and value $v$. 
\begin{equation}
    V_i = \{\tau_i, g(\tau_i), c(\tau_i), v_i \}
    \label{eq:structure}
\end{equation}
We use the shorthand notations $g(V_i), c(V_i), v(V_i)$, and $\tau(V_i)$ to refer to the quantities associated with node $V_i$.
The decomposition into gain, cost, and value allows for separate updates of these quantities, enabling efficient maintenance of large trees by caching or storing values until they change.
In addition, the following criteria need to be met:
the gain $g(V_i)$ can be any function that depends only on the pose at the end of the trajectory $\tau(V_i)$, such as the number of observable voxels.
The gains are assumed to be mutually independent, not accounting for e.g. frustum overlap.
The cost $c(V_i)$ is required to be an intrinsic property of $V_i$, for example the length of the associated trajectory.
The value $v(V_i)=f(g(V_1), \dots, g(V_n), c(V_1), \dots, c(V_n))$ fuses cached gains $g$ and costs $c$ into the total utility of a node.

When searching for the `next-best' node, all nodes are considered, and the one containing the highest value $v$ in its sub-tree is used.
Any definition of $g$, $c$, and $v$ adhering to these conditions is admissible, allowing the algorithm to maximize an application-specific objective.
Specific choices of $g$, $c$, and $v$ are further discussed in section \ref{sec:cost_utility}.

\subsection{Tree Expansion}
To expand the tree, a new viewpoint is sampled. 
Although any sampling procedure is allowed, we use a two stage approach to guarantee sufficient local coverage and reachability of viewpoints.
If less than $n_{local}$ viewpoints are within a distance $r_{local}$ of the robot, the new point is sampled from a sphere with radius $r_{local}$. 
Otherwise we use the method of \cite{nbvp_bircher} to sample a global point.
To ensure reachability and encourage viewpoints near exploration boundaries, a linear path is extended from the nearest neighbor node in the tree towards the sampled point, until either a maximum edge length $l_{max}$ or a safe distance to inaccessible space is reached. 
Because the gain only depends on the viewpoint, it can now be computed and only needs to be updated when the map is modified.

To add the new node to the tree, we perform an operation similar to the RRT* algorithm \cite{rrt_star}.
Instead of connecting the viewpoint to its nearest neighbor node, as done in standard RRT, we try to connect the viewpoint to all nodes within $l_{max}$ and select the parent node that maximizes the value of the new node.
The value can be checked efficiently, as previously computed gains and costs do not change.
Additionally, the remaining neighbors are rewired to the newly added node if this increases their value, thereby refining the current tree.

\subsection{Tree Updating}
Similar to \cite{rt_rrt_star}, when a new node is executed, that node becomes the new root of the tree, and its gain, cost, and value are set to zero.
To prevent loss of the non-executed branches, they are rewired, if possible, to nearby nodes that remain connected to the root.
Opposed to \cite{rt_rrt_star}, due to the high cost of the gain computation, we keep the entire tree alive. If rewiring is not successful, the old root is re-sampled with its start pose as viewpoint to guarantee that nodes can be rewired unless the map has changed to render nodes inaccessible.
To update the tree, we make use of the previously introduced node structure \eqref{eq:structure}.
Since the costs are intrinsic to each node, they do not need to be updated. 
The gains only need to be updated in areas where the map has changed. 
We model this behavior by only updating nodes within a maximum distance $r_{update}$ of the robot's current position and whose gain has not yet reached zero. 
Because the value is a function of the cached gains and costs, it can be updated efficiently in a single pass through the tree.
As the value of a node can change drastically during updating, the tree is re-optimized by performing a rewiring step for every node in a breadth-first manner.
The maximum tree size is thus implicitly governed by the updating computational cost and the computational resources available.
As more segments are only sampled in the time remaining after updating, the algorithm aims to expand the largest tree that can be kept fully updated with the computational resources available.


\section{COST-UTILITY FORMULATIONS}
\label{sec:cost_utility}
The presented algorithm shapes the tree to maximize the value $v$ and is thus strongly governed by the choice of $g(V_i)$, $c(V_i)$, and $v(V_i)$.
In the following, we present a gain formulation for the application of accurate TSDF-based 3D reconstruction under uncertainty in sensing and state estimation.
We address the problem of how to combine general information gains and costs into the total utility $v$ and present a parameter-free efficiency inspired value formulation.
\subsection{TSDF-based 3D Reconstruction Gain}
A TSDF map typically consists of a voxel grid, where each voxel $\mathbf{m}$ contains a distance $d(\mathbf{m})$ and a weight $w(\mathbf{m})$. 
This representation has the advantage that surface models can be directly created from the map, e.g. using the marching cubes algorithm \cite{marching_cubes}.
In addition, the weights can be treated as an uncertainty measure.

Commonly used 3D sensors, such as the Microsoft Kinect or Intel Realsense, have measurement noise that scales approximately quadratically with depth $z$ \cite{kinect_noise,realsense_noise}. Thus, we use the quadratic weight proposed in \cite{voxblox} to minimize sensing errors.
\begin{equation}
w_{in}(\mathbf{m}) = z^{-2}.
\label{eq:input_weight}
\end{equation}
The error introduced by uncertainty in the state estimation is more difficult to estimate, since it is highly dependent on the shape of the surface. 
The risk of mapping a point onto an unrelated voxel, introducing potentially large errors, increases greatly with the distance between the measured point and its true location. 
Assuming an unknown offset in the robot pose, this distance is typically dominated by the orientation-related error, which scales linearly w.r.t. $z$. 
To minimize this risk we employ the same quadratic penalty as a heuristic.
Utilizing the following update rule for the map distance $d(\mathbf{m})$,
\begin{equation}
    d_{new}(\mathbf{m}) = \frac{w(\mathbf{m})d(\mathbf{m}) + w_{in}(\mathbf{m})d_{in}(\mathbf{m})}{w(\mathbf{m}) + w_{in}(\mathbf{m})},
    \label{eq:map_update}
\end{equation}
we propose to use the impact a view has on the map as the information gain. 
Combining \eqref{eq:input_weight} and \eqref{eq:map_update} and accounting for multiple rays traversing a voxel, the impact factor $\eta(\mathbf{m})$ is computed as:
\begin{equation}
    \eta(\mathbf{m}) = \dfrac{1}{1 + z^2 w(\mathbf{m})/N_{rays}(\mathbf{m})},
\end{equation}
where the number of rays intersecting a voxel $N_{rays}(\mathbf{m})$ depends on the employed camera and is usually also proportional to $z^{-2}$.

Similar to \cite{Yoder2016}, we focus on surfaces, evaluating the impact factor only for observed surfaces $\mathbb{S}$ and unknown voxels near observed surfaces $\mathbb{U}$. The resulting information gain $g_{rec}(\xi)$ of a viewpoint $\xi$ is thus given by,
\begin{equation}
    g_{rec}(\xi) = \sum_{\mathbf{m}\in\text{Vis}(\xi)}  
    \begin{dcases} 
    1, & \text{if } \mathbf{m} \in \mathbb{U},\\
    \dfrac{\eta(\mathbf{m})-\eta_{min}}{1-\eta_{min}}, & {\text{if } \mathbf{m} \in \mathbb{S} \text{ and } \atop \eta(\mathbf{m}) > \eta_{min}},\\ 
    0, & \text{otherwise},
    \end{dcases}
    \label{eq:g_rec}
\end{equation}
where $\text{Vis}(\xi)$ is the set of voxels visible from $\xi$ and $\eta_{min}$ is a cut-off value for low impact factors.
The resulting gain is simultaneously a quality and an efficiency measure. Intuitively this gain favors viewpoints that observe new areas and improves previously mapped voxels.

\subsection{Global Normalization Value}

Oftentimes, part of a task's objective is to execute it as quickly as possible. Therefore, we use the expected execution time $t$ as the cost of a node $V_i$, i.e.:
\begin{equation}
    c(V_i) = t_{end}(\tau(V_i))-t_{start}(\tau(V_i))
    \label{eq:time_cost}
\end{equation}
The value function $v(V_i)$ addresses the issue of how to connect gains and costs to arrive at a measure of total utility.
Commonly used value formulations such as exponential $v_{exp}(V_i)$ \cite{nbvp_bircher, song_3d_rec, Selin_nbv_fron} or linear $v_{lin}(V_i)$ \cite{Yoder2016,aut_expl_comparison, gmm_multirobot_expl} penalties,
\begin{align}
    v_{exp}(V_i) & = v(\text{parent}(V_i)) + g(V_i)\exp{(-\lambda c(V_i))} \label{eq:exp_cost},\\
    v_{lin}(V_i) & = v(\text{parent}(V_i)) + g(V_i) -\alpha c(V_i) \label{eq:lin_cost},
\end{align} 
have several disadvantages. $v_{exp}$ is strictly increasing, favoring long sub-trees even when subsequently executed nodes might be sub-optimal.
Furthermore, both require careful tuning of the parameters $\lambda$ and $\alpha$.
To address these two issues, we use the notion of efficiency, i.e. the accumulated gain per cost, as a central idea for the value. 
Instead of greedily maximizing the efficiency of each node, we account for the global context of each node by defining the global normalization value $v_{GN}(V_i)$ as,
\begin{equation}
    v_{GN}(V_i) =
        \max_{V_k \in \text{subtree}(V_i)}
        \frac{
            \sum_{V_l \in \mathcal{R}(V_k)}g(V_l)
        }{
            \sum_{V_l \in \mathcal{R}(V_k)}c(V_l)
    },
    \label{eq:gnc_cost}
\end{equation}
where $\text{subtree}(V_i)$ indicates the tree originating from and including $V_i$ and $\mathcal{R}(V_i)$ is the sequence of nodes connecting $V_i$ to the root of the tree.


\section{IMPLEMENTATION DETAILS}
The presented algorithm and utility formulations are implemented using the \textit{mav\_active\_3d\_planning} package\footnotemark[1], a modular IPP design framework we open-source with the algorithm.
While Voxblox \cite{voxblox} is used as the map representation any other TSDF implementation could be used.
We furthermore perform yaw optimization as presented in \cite{history_nbvp} to compute gains.
The entire yaw spectrum is divided into sections, spanning 30$^\circ$ in our case, and the gain is computed for each section.
The yaw maximizing the total gain of a node is then greedily selected.
We use a constant acceleration model for position and yaw to estimate the execution time of future trajectories and polynomial trajectory optimization \cite{richter2016polynomial} to the executed paths.

\subsection{Iterative Ray Casting}
Because the majority of computation time is spent evaluating the gain of a viewpoint, an efficient ray casting method, which we call iterative ray casting, is used. 
The initial ray spacing is reduced to match the voxel size $s$ at maximum ray length. 
The cast rays add all traversed voxels to a set of visible voxels. 
Subsequent rays only start where their lateral distance to the previous rays is $s$.
This reduces redundant voxel checks in areas of high ray density, speeding up computation by an additional 233\% in our experiments, while still detecting all voxels.
As shown in \cite{Selin_nbv_fron, safe_local_exploration}, computation can be further sped up by approximating the gain computation. 
When multiplying $s_{ray}$ with a sub-sampling factor $f_{sub}$, the density of detected voxels per volume and per surface decreases proportional to $f_{sub}^2$. 
Because the planner selects nodes based on a relative value, no further compensation is required.

\begin{figure}[t]
     \centering
     \includegraphics[width=\columnwidth]{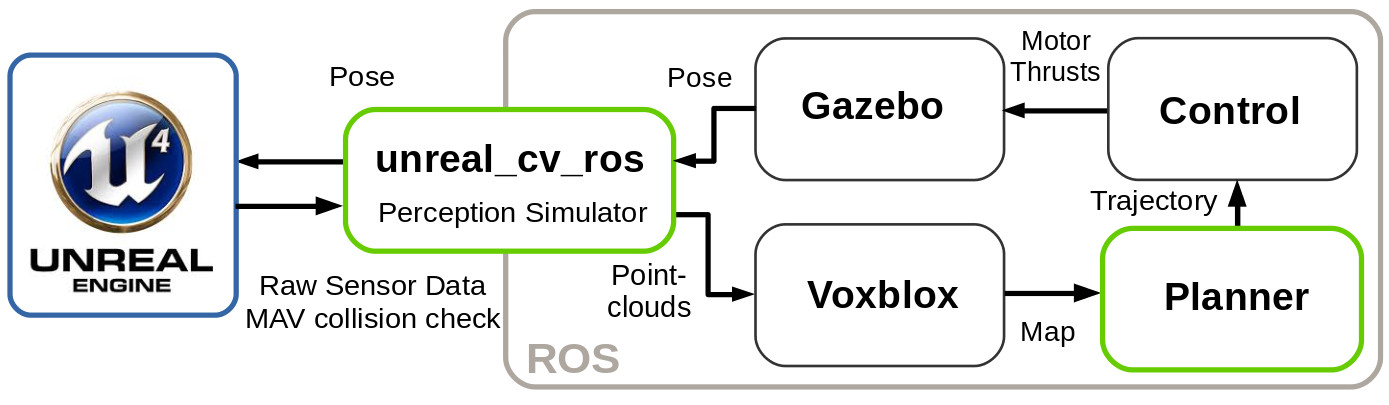}
     \caption{Simulation environment overview.}
     \label{fig:simulation}
\end{figure}

\begin{table}[t]
    \centering
    \caption{Parameters used throughout all experiments.}

    \begin{tabular}{llll}
        \toprule
        Max. Velocity & \SI{1}{\meter\per\second} & Max. Acceleration & \SI{1}{\meter\per\square\second} \\
        Max. Yaw Rate & \SI{\pi/2}{\radian\per\second} & Collision Radius & \SI{1.2}{\meter} \\
        Camera FOV & \SI{90}{\degree}$\times$\SI{73.7}{\degree} & Camera Range & \SI{5}{\meter} \\
        Voxel Size & \SI{0.1}{\meter} & Edge Length $l_{max}$ & \SI{1.5}{\meter} \\
        \bottomrule
    \end{tabular}

    \label{tab:system_constraints}
\end{table}

\begin{table}[t]
    \centering
    \caption{Parameters of the proposed planner. }

    \begin{tabular}{lll}
        \toprule
        Local Sampling Count & $n_{local}$ & \num{10} \\
        Local Sampling Radius & $r_{local}$ & \SI{1.5}{\meter} \\
        Updating Radius & $r_{update}$ & \SI{3}{\meter} \\
        Minimum Impact Factor & $\eta_{min}$ & \num{0.01} \\
        Ray Sub-sampling Factor & $f_{sub}$ & \num{3.0} \\
        \bottomrule
    \end{tabular}
    
    \label{tab:planner_params}
\end{table}

\fboxsep=0mm
\fboxrule=0.75pt
\begin{figure*}
     \centering
     \begin{subfigure}[c]{0.32\textwidth}
        \fcolorbox{gray}{white}{\includegraphics[width=0.98\linewidth]{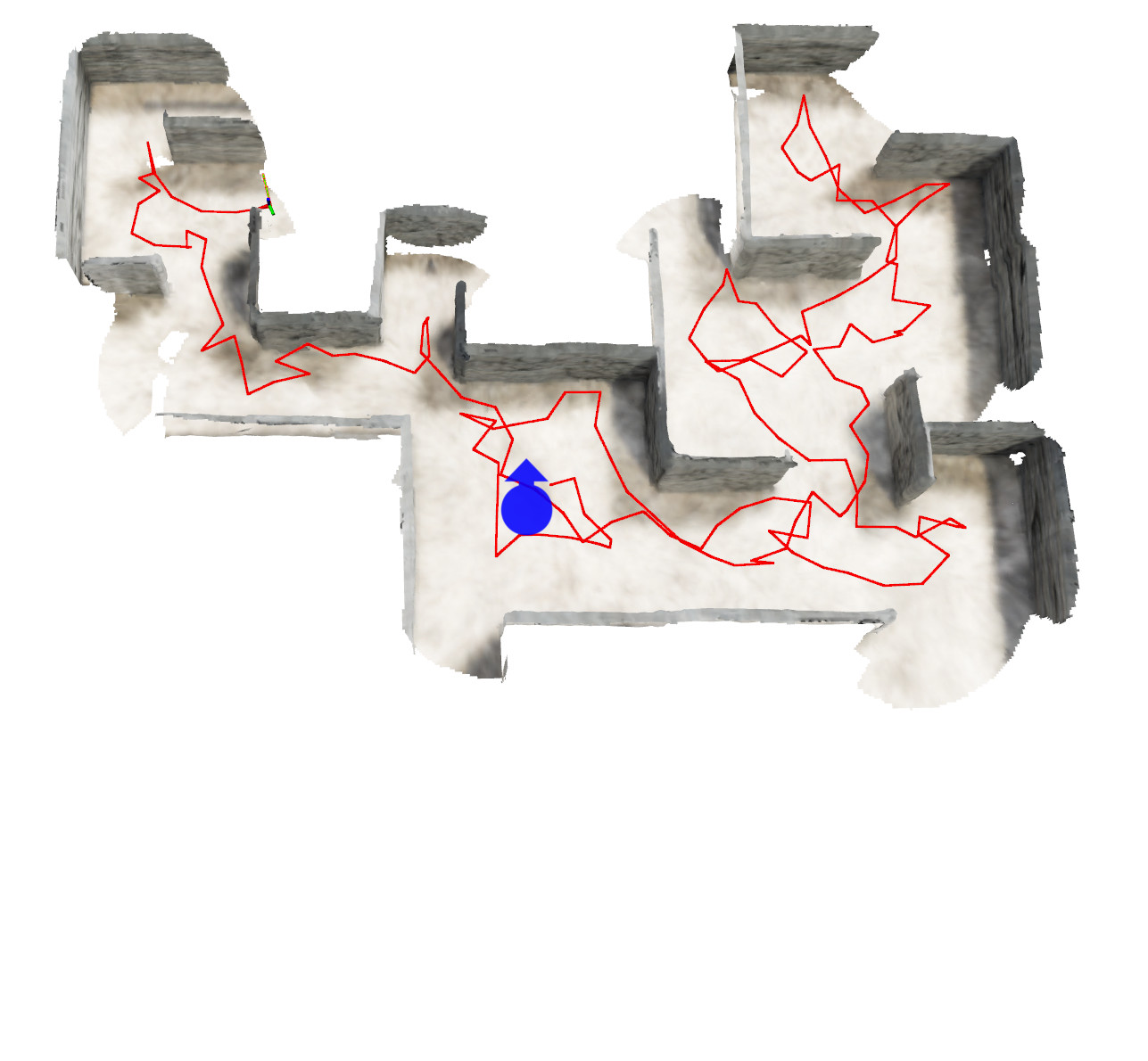}}
         \caption{RH-NBVP \cite{nbvp_bircher}}
         \label{fig:exploration_map_1}
     \end{subfigure}
     \begin{subfigure}[c]{0.32\textwidth}
        \fcolorbox{gray}{white}{\includegraphics[width=0.98\linewidth]{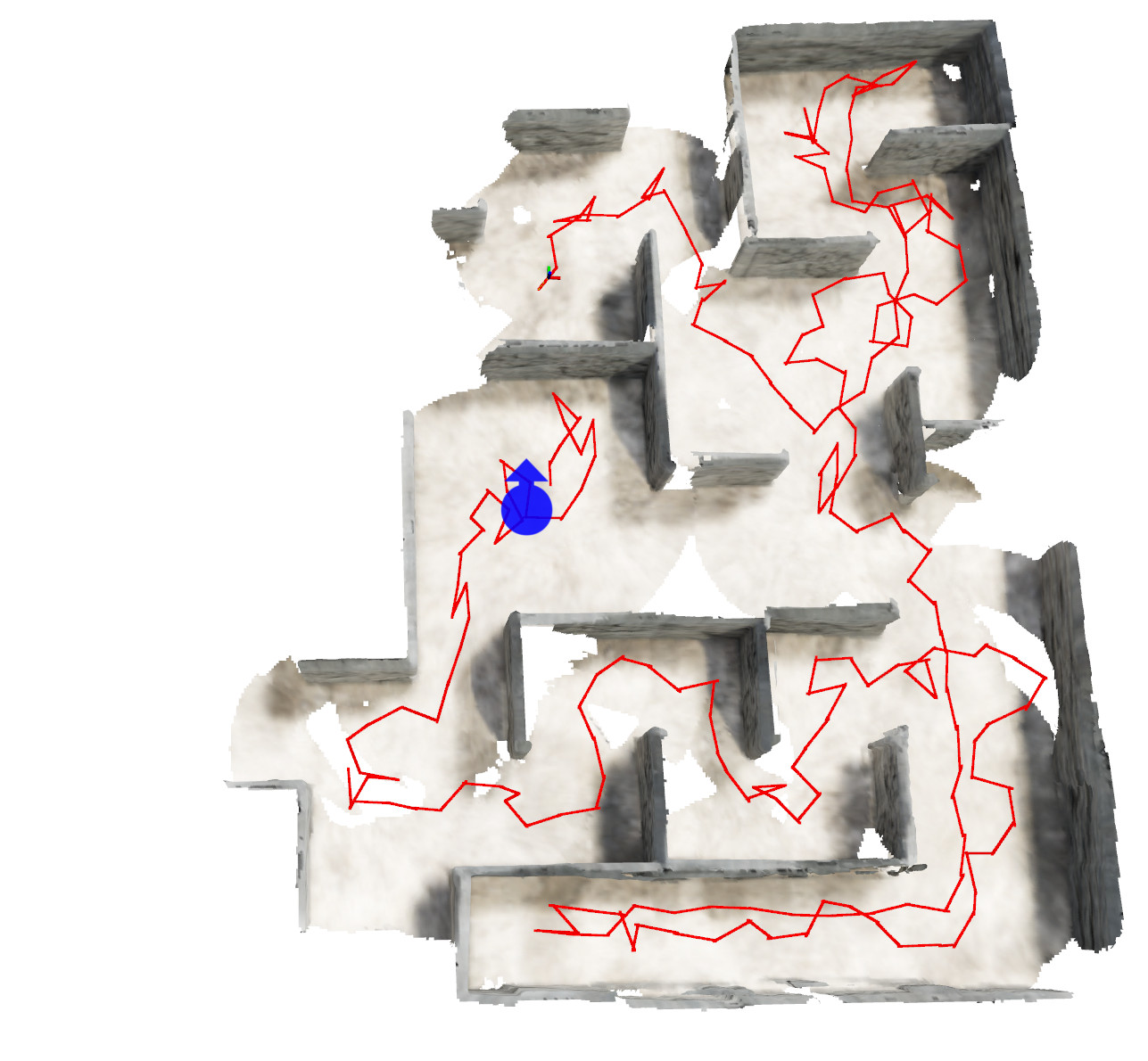}}
         \caption{AEP \cite{Selin_nbv_fron}}
         \label{fig:exploration_map_2}
     \end{subfigure} 
     \begin{subfigure}[c]{0.32\textwidth}
        \fcolorbox{gray}{white}{\includegraphics[width=0.98\linewidth]{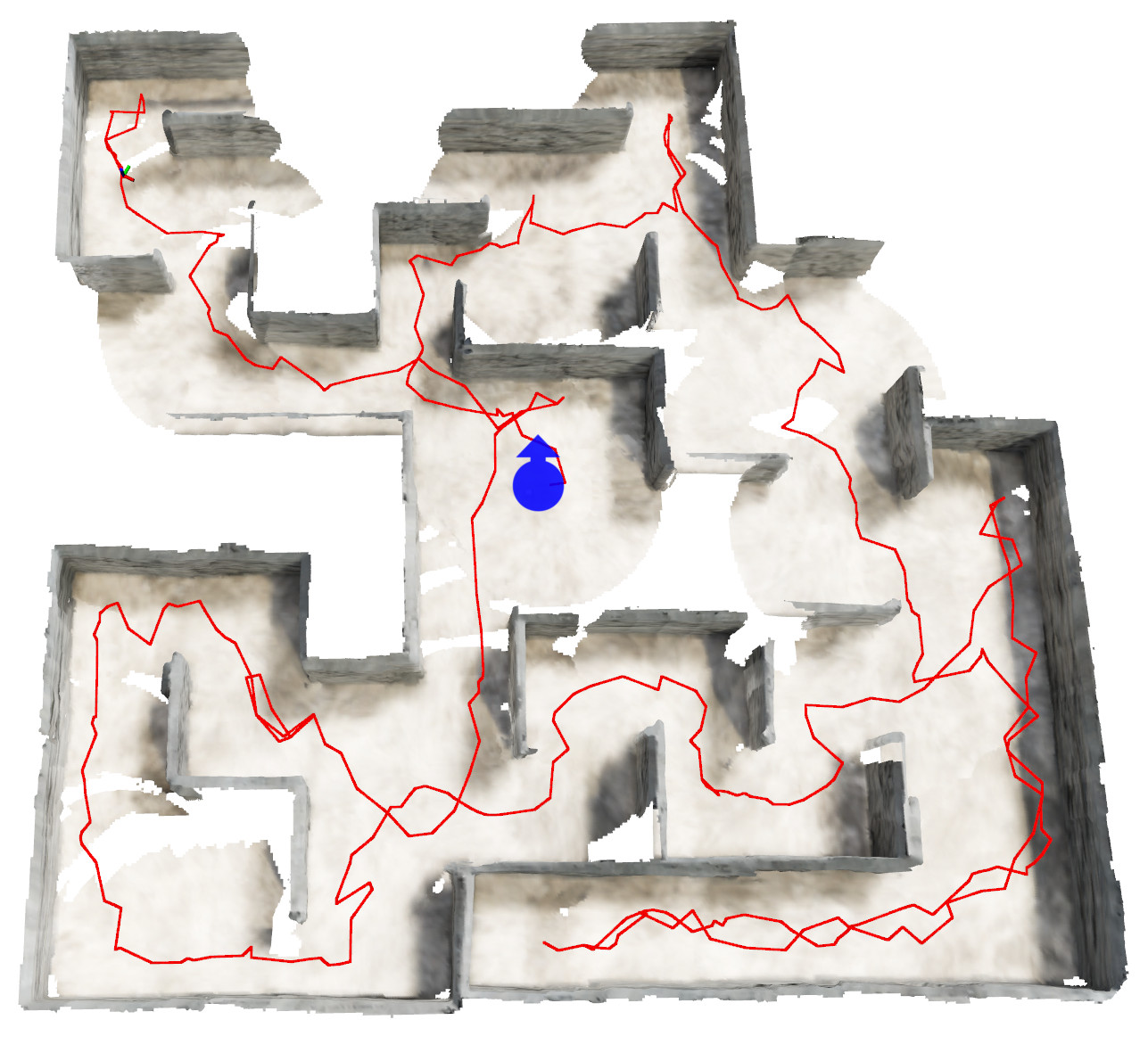}}
         \caption{Ours}
         \label{fig:exploration_map_3}
    \end{subfigure}
    \caption{Executed paths after 15 minutes of autonomous exploration in the maze scenario, starting from the blue pose. Although all three methods sample linear paths, our method's capability to refine the tree to maximize global utility produces less jagged paths. This results in a larger area being explored in the same time while covering a distance similar to AEP.}
    \label{fig:exploration_map}
\end{figure*}

\begin{figure*}
     \centering
     \begin{subfigure}[c]{0.32\textwidth}
        \includegraphics[width=\textwidth]{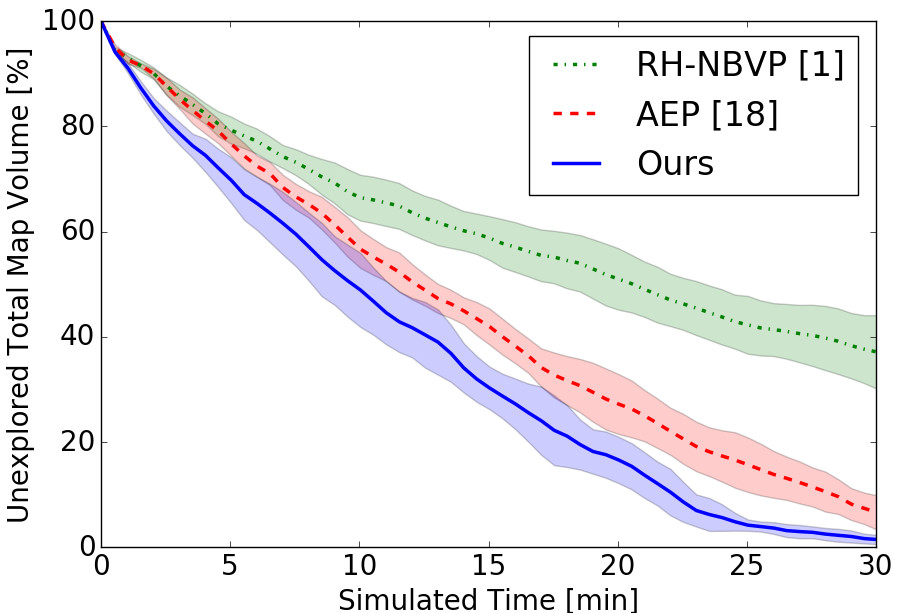}
        \caption{Maze Exploration Rate}
        \label{fig:exploration}
    \end{subfigure}
     \begin{subfigure}[c]{0.32\textwidth}
         \includegraphics[width=\textwidth]{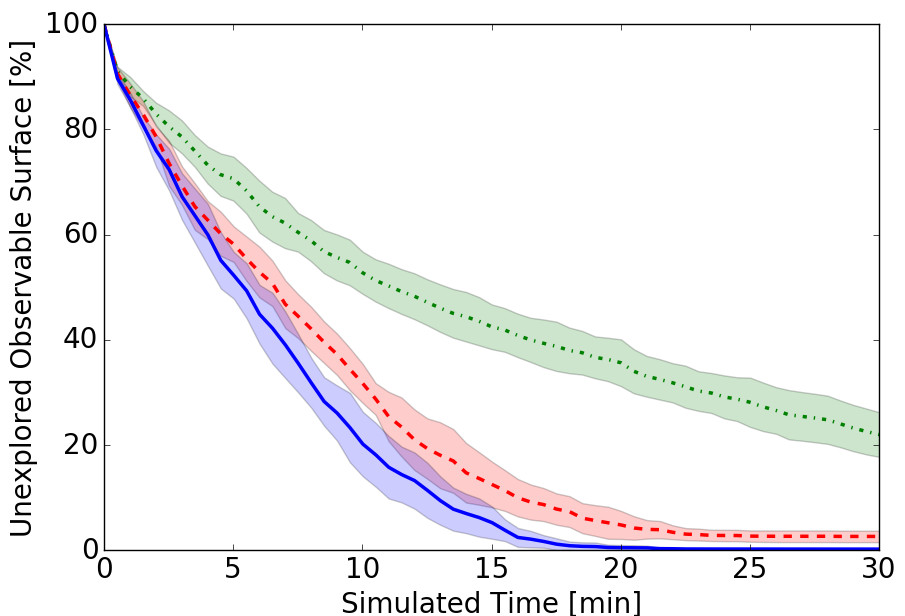}
         \caption{Building Exploration Rate}
         \label{fig:city_graphs_a}
     \end{subfigure}
     \begin{subfigure}[c]{0.32\textwidth}
         \includegraphics[width=\textwidth]{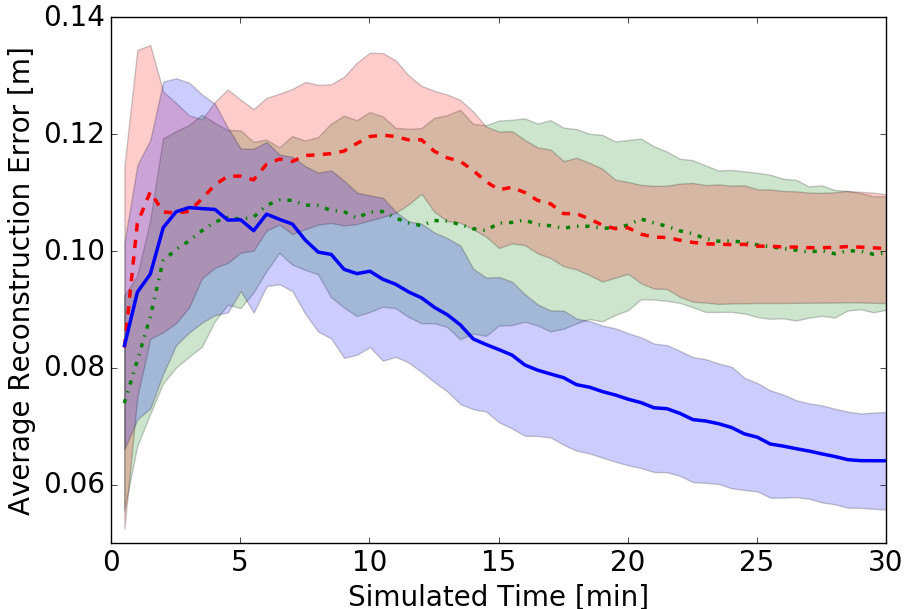}
         \caption{Building Reconstruction Error}
         \label{fig:city_graphs_b}
     \end{subfigure} 
    \caption{Exploration progress for the maze environment (left) and building (center). Average reconstruction error on the building environment (right). Mean and standard deviation over 10 experiments are shown.}
    \label{fig:city_graphs}
\end{figure*}

\section{SIMULATION ENVIRONMENT}
\label{sec:sim_env}
Active path planning tasks require a simulator rather than a dataset, thus a realistic simulation environment is advantageous. 
We base our simulated world on the Unreal Engine\footnote{https://www.unrealengine.com/en-US/}, a game engine capable of rendering photo-realistic scenes.
The behavior of a MAV is simulated using the Robot Operating System\footnote{https://www.ros.org/} (ROS).
An overview of the simulator is given in Fig.~\ref{fig:simulation}. 
We use voxblox \cite{voxblox} as the map representation, the control hierarchy presented in \cite{kamelmpc2016} is used for trajectory tracking, and the Gazebo-based RotorS simulator \cite{rotors_gazebo} accurately models the MAV's physics.
The perception interface\footnotemark[2] is based on the UnrealCV computer vision plugin \cite{unrealcv} and is optimized for robotic applications such that it runs in real time at \SI{3}{\hertz} and checks for collisions in the simulated world. 
To account for sensing uncertainty, a Gaussian depth error with quadratic scaling \cite{kinect_noise, realsense_noise} is added to the ground truth point cloud, i.e.:
\begin{equation}
    z_{sim} = z + \mathcal{N}(\mu(z),\sigma(z)), \, \mu(z) = \sigma(z) = 0.0024 z^2.
    \label{eq:sensor_noise}
\end{equation}
Uncertainty in state estimation is modeled by a bounded random walk, introducing an error in position within a sphere of radius \SI{5}{\centi\meter}, roll and pitch within $\pm$~\SI{1.5}{\degree} and yaw within $\pm$~\SI{5}{\degree}. The values are chosen as best estimates of errors achievable by state of the art systems such as RTK-GPS.

Exploration is evaluated in a maze scenario of \SI{40}{\meter}$\times$\SI{40}{\meter}$\times$\SI{3}{\meter}, while
for 3D reconstruction, an urban environment of size \SI{40}{\meter}$\times$\SI{40}{\meter}$\times$\SI{12}{\meter} is used.
Particularly challenging areas are the trees occluding the left side, narrow passages on the right, and the filigree geometry encountered on the roof windows.
We make the simulator and scenarios available for benchmarking and future research.


\section{EXPERIMENTS}
We compare our method to the receding horizon NBV planner (RH-NBVP) \cite{nbvp_bircher} and the more recent autonomous exploration planner (AEP) \cite{Selin_nbv_fron}. 
Both methods were shown to perform well in indoor exploration as well as volumetric reconstruction of a bridge and power plant building, respectively.
For all experiments, identical system constraints, tree edge length, and  TSDF update weights \cite{voxblox} were used (Table~\ref{tab:system_constraints}). 
Additional parameters of our method are listed in Table~\ref{tab:planner_params}.

Two independent metrics are used to evaluate the performance of the methods: the exploration ratio, which captures the percentage of the environment observed, and average reconstruction error, which measures the average error between the reconstructed surface and ground truth.
The exploration ratio thus indicates the completeness of the reconstruction and the reconstruction error the average quality of the obtained model.
Due to the stochastic nature of the planners all experiments are repeated 10 times and the means and standard deviations are reported.

\subsection{Volumetric Indoor Exploration}
We evaluate the ability of our algorithm to generate efficient paths in the application of volumetric exploration.
Therefore, `unknown volume' as used by RH-NBVP and AEP is substituted as gain $g(V_i)$.
Fig.~\ref{fig:exploration} shows the exploration rate of each method in the maze scenario. 
Even though different paths were taken in different experiments, our method achieves full exploration of the observable map volume after 25 minutes in all runs, showing its global coverage capabilities.
A visual analysis is given in Fig.~\ref{fig:exploration_map}, showing the executed path (starting from the center) after 15 minutes of exploration. 
RH-NBVP suffers from several issues, including: long computation times when stuck in a dead end, spending a lot of time turning on the spot, and frequent changes of the exploration site. Consequently RH-NBVP travels only \SI{210}{\meter}, covering 60\% of the environment.
By contrast AEP and our method keep constantly moving, covering a distance of \SI{297}{\meter} and \SI{330}{\meter}, respectively.
As our proposed method maintains and refines a trajectory tree it has access to more and higher quality paths then methods such as AEP which rebuild a tree in every iteration. This results in the selection of smoother and more informative paths. Thus even when using the same linear path representation and gain formulation as the other methods our proposed method achieves a higher convergence rate and final map quality.


\begin{figure*}[bt]
\centering
\begin{adjustbox}{width=0.95\textwidth} 
    \setlength{\tabcolsep}{2pt}
    \begin{tabular}{ccc}
        RH-NBVP \cite{nbvp_bircher} & AEP \cite{Selin_nbv_fron} & Ours \\
        \includegraphics[width=0.33\linewidth]{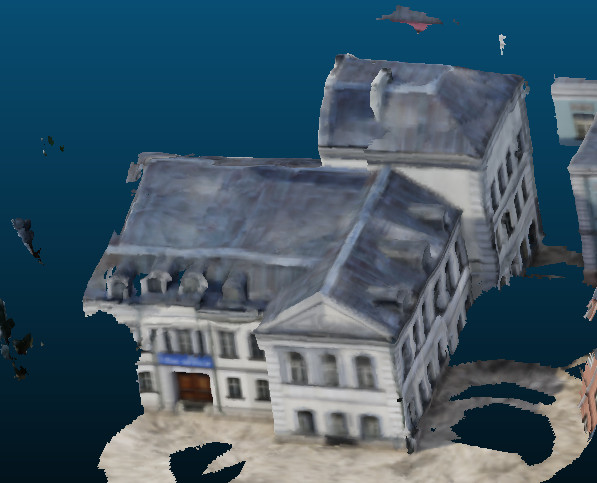} &
        \includegraphics[width=0.33\linewidth]{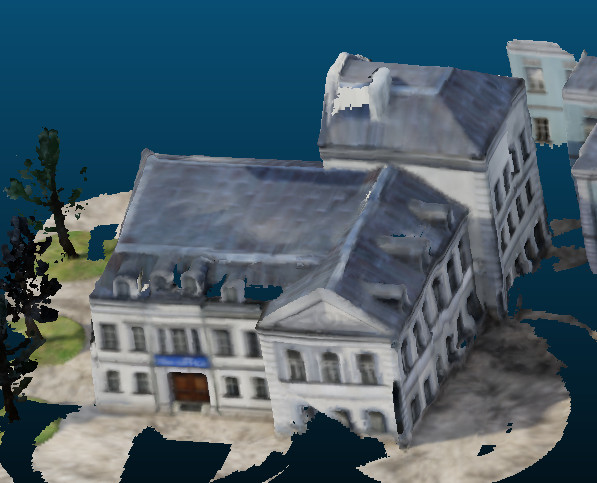} &
        \includegraphics[width=0.33\linewidth]{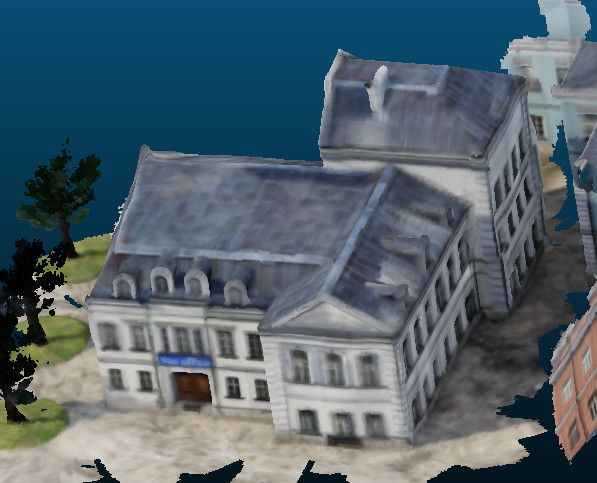} \\
        \includegraphics[width=0.33\linewidth]{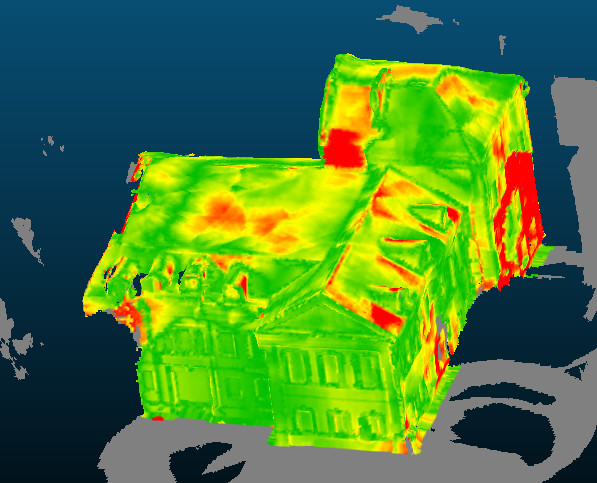} &
        \includegraphics[width=0.33\linewidth]{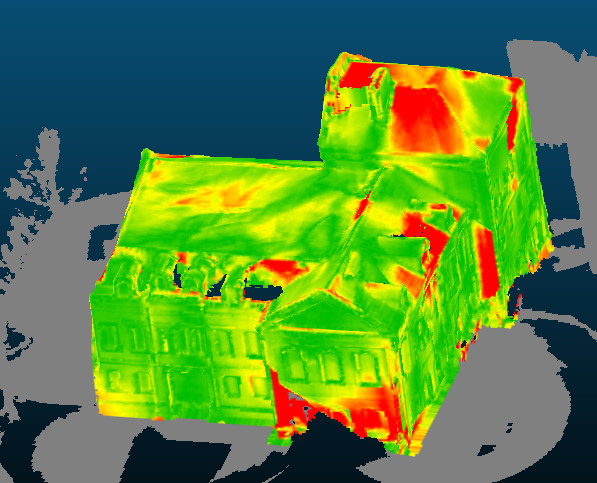} &
        \includegraphics[width=0.33\linewidth]{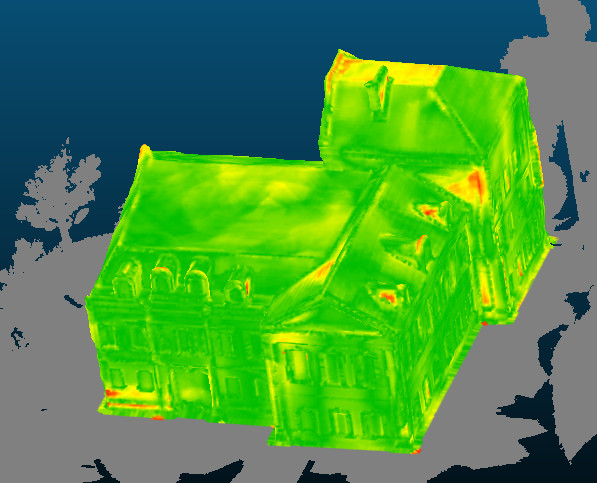} \\
        \multicolumn{3}{c}{\includegraphics[width=0.65\textwidth]{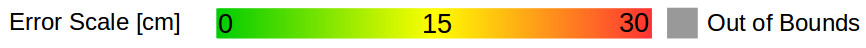}} \\
    \end{tabular}
\end{adjustbox}
\caption{Obtained 3D reconstruction (top row) and reconstruction error (bottom row) after 30 minutes. Our method is able to fully reconstruct the building without leaving any holes in the mesh. Error peaks are suppressed by focusing on areas of high expected error, resulting in higher overall accuracy and a more homogeneous error distribution.}
\label{fig:city_qualitative}
\end{figure*}

\subsection{3D Reconstruction}
The objective in the urban scenario is to accurately reconstruct the central building. 
To allow for a fair comparison between volumetric and surface-based gains, a tight region of interest bounding box around the target is used. 
The exploration rates and average reconstruction errors are shown in Fig.~\ref{fig:city_graphs_a} and Fig.~\ref{fig:city_graphs_b}.
By sampling global points only, RH-NBVP has difficulties exploring narrow and occluded areas.
Both \cite{nbvp_bircher} and \cite{Selin_nbv_fron} put little emphasis on small volumes, especially when they are far away, resulting in holes remaining unexplored.
While AEP explores the map faster than RH-NBVP, some occlusions escape its attention, leading to termination with unexplored holes still in the map.
Expanding a large tree, our method is able to fully explore the building in all of the experiments within 20 minutes.
Our gain formulation initially focuses on global surface exploration, before focusing on areas of low map weight, i.e. low expected reconstruction quality, when no more unobserved areas are present.
This behavior is apparent after 10 minutes, when exploration slows down, the error is further reduced.
 This not only increases the overall accuracy but also results in a more homogeneous error distribution with lower maximum error.
 Due to the quality unaware gain, similar average reconstruction errors are achieved for \cite{nbvp_bircher} and \cite{Selin_nbv_fron}.
The difference in behavior and resulting difference of \SI{4}{\centi\meter} in average reconstruction error is visualized in Fig.~\ref{fig:city_qualitative}.

\subsection{Importance of Sampling and Gain Formulation}

\begin{table}[bt]
    \centering
    \caption{City environment ablation study of various components.}

    \begin{tabular}{lrrr}
        \toprule
            \parbox{0.2\columnwidth}{Method} &
            \parbox{0.18\columnwidth}{\centering Exploration 80\% [min]} &
            \parbox{0.18\columnwidth}{\centering Exploration \SI{30}{\minute} [\%]} &
            \parbox{0.2\columnwidth}{\centering Reconstruction Error [cm]} \\
        \midrule
        Our, RRT$^*$, $g_{rec}$ & $10.7 \pm 1.3$ & $\mathbf{99.8 \pm 0.1}$ & $ \mathbf{6.4 \pm 0.8}$ \\
        Our, RRT$^*$, $g_{uv}$  & $\mathbf{9.7 \pm 1.2}$ & $\mathbf{99.8 \pm 0.1}$ & $ 8.9 \pm 0.7$ \\
        Our, RRT, $g_{rec}$     & $ 13.8 \pm 2.6$ & $99.6 \pm 0.4$ & $ 8.0 \pm 0.5$ \\
        \cite{nbvp_bircher}, RRT, $g_{rec}$
                                & $17.5 \pm 2.8$ & $93.9 \pm 4.7$ & $ 8.2 \pm 1.0$ \\
        \cite{nbvp_bircher}, RRT, $g_{uv}$
                                & $22.9 \pm 2.6$ & $89.8 \pm 5.4$ & $10.2 \pm 1.2$ \\
        \bottomrule
    \end{tabular}
    
    \label{tab:ablation}
\end{table}

To investigate the impact of the components of our proposed method we evaluate different combinations of gain and sampling strategies. Methods denoted with \emph{Our} keep the exploration tree alive, while those denoted with \emph{\cite{nbvp_bircher}} discard non-executed branches. Both RRT and RRT$^*$ are explored as sampling schemes using both the `unknown volume' $g_{uv}$ and our proposed $g_{rec}$ gain formulations. The results in Table~\ref{tab:ablation} show that retaining the tree significantly improves global coverage. RRT$^*$ rewiring improves the exploration speed while providing an improvement in reconstruction quality with identical gain formulations. Looking at the results for the different gains we can see that the proposed gain $g_{rec}$ improves over the simpler $g_{uv}$ gain independent of the sampling method employed. Finally, it is the combination of retaining the tree structure, continuously refining paths in an RRT$^*$ manner and the novel gain formulation that provides results not achievable with other combinations.


\begin{table}[bt]
    \centering
    \caption{Impact of different information gains.}

    \begin{tabular}{lrr}
        \toprule
            \parbox{0.25\columnwidth}{Information Gain} &
            \parbox{0.25\columnwidth}{\centering Exploration\\ 95\% [min]} &
            \parbox{0.25\columnwidth}{\centering Reconstruction\\ Error [cm]} \\
        \midrule
        Unknown Volume                      & $15.0 \pm 1.9$ & $8.9 \pm 0.7$  \\
        Surface Frontiers \cite{Yoder2016}  & $16.2 \pm 1.5$ & $7.8 \pm 0.8$ \\
        Voxel Confidence \cite{song_3d_rec} & $15.6 \pm 1.8$ & $8.5 \pm 1.1$ \\
        Voxel Impact (ours)                 & $\mathbf{14.7 \pm 1.7}$ & $\mathbf{6.4 \pm 0.8}$ \\
        \bottomrule
    \end{tabular}

    \label{tab:gain}
\end{table}

\subsection{Gain and Value Formulation}
Because our algorithm is strongly governed by the objective function, we investigate the influence of different gains, costs, and value formulations in the city building scenario. 
In a first experiment, we evaluate the effect of the gain $g(V_i)$.
The cost $c(V_i)$ is fixed to the execution time \eqref{eq:time_cost} and the value $v(V_i)$ fixed to $v_{GN}(V_i)$ \eqref{eq:gnc_cost}.
We compare our proposed voxel impact gain \eqref{eq:g_rec} against the number of unknown voxels, surface frontiers \cite{Yoder2016}, and confidence voxels \cite{song_3d_rec}. 
For \cite{Yoder2016}, a maximum number of points $t_m=50$ and safety distance $d_s = \SI{1}{\meter}$ is used. 
For the voxel confidence, voxels count if they are unobserved or if their normalized weight $\bar{w}(\mathbf{m}) < \theta_{conf}=0.4$. 
The mean and standard deviation of 10 experiments are reported in Table~\ref{tab:gain}.
While the exploration times are comparable, the volumetric gains heavily depend on a carefully chosen region of interest to attain quick exploration.
While most gain formulations trade off exploration speed for increased quality, our gain quickly explores the building and then further refines areas of high uncertainty, resulting in the lowest final reconstruction error.
Although \cite{song_3d_rec} also revisits areas of low map weight, the selected viewpoints are of similar quality as before and do not translate into a significant increase in quality.

\begin{table}[bt]
    \centering
    \caption{Impact of different cost and value formulations.}
    
    \begin{tabular}{lrr}
        \toprule
            \parbox{0.25\columnwidth}{Value Formulation} &
            \parbox{0.25\columnwidth}{\centering Exploration \\ \SI{30}{\minute} [\%]} &
            \parbox{0.25\columnwidth}{\centering Reconstruction \\ Error [cm]} \\
        \midrule
        Linear penalty      & $76.4 \pm 6.5$ & $8.9 \pm 1.4$ \\
        Exponential penalty & $96.6 \pm 5.3$ & $8.0 \pm 0.8$ \\
        Global Norm. (ours) & $\mathbf{99.8 \pm 0.1}$ & $\mathbf{6.4 \pm 0.8}$ \\
        \bottomrule
    \end{tabular}
    
    \label{tab:cost}
\end{table}

Next, we fix the gain $g(V_i)$ to the proposed voxel impact \eqref{eq:g_rec} and compare the proposed value function $v_{GN}(V_i)$ \eqref{eq:gnc_cost} against a linear penalty \eqref{eq:lin_cost} on the execution time and an exponential penalty \eqref{eq:exp_cost} on the accumulated path length as cost and value formulation. 
The results obtained with $\alpha=3.0$ and $\lambda=0.5$ are given in Table~\ref{tab:cost}.
The importance of the cost and value function choice for global planning is apparent as neither penalty is able to reliably reconstruct the building.
By construction both linear and exponential penalties favor sub-optimal viewpoints that have low cost which results in higher reconstruction error values. For the linear penalty, the obtained error increases by 39\% compared to our proposed method and matches the error of the quality agnostic `unknown volume' gain of Table~\ref{tab:gain}, thus completely negating the advantages of the employed gain.
Although careful tuning of these penalties could improve their performance, the reliance on parameter tuning is a considerable disadvantage -- especially when compared to our proposed gain, which requires no parameter tuning.


\subsection{Robot Experiment}

We deploy our proposed method as well as RH-NBVP \cite{nbvp_bircher} on a real MAV. Fig.~\ref{fig:real_room} shows the reconstructed environment after \SI{3}{\minute} flight time. The MAV is equipped with a Realsense D435 sensor and uses the same parameters as the simulation experiments. State information was provided by a Vicon system to remove compounding errors induced by state estimation drift. We can see the qualitative difference between the trajectories and the final reconstructed scene in Fig.~\ref{fig:real_room}. Our proposed method traverses a larger part of the environment and obtains a reconstruction with no holes and straight walls. By comparison, RH-NBVP remains in a small area, failing to cover certain viewpoints, which results in the reconstruction containing holes and non-straight walls. This difference in behavior and reconstruction quality is explained by our cost function reasoning about the sensor's noise, which encourages moving closer to surfaces as opposed to simply ensuring observing voxels.


\section{CONCLUSIONS}

In this paper, we presented a new sampling-based online informative path planning algorithm.
Our approach obtains global coverage by continuously expanding a single trajectory tree, allowing the algorithm to maximize a single objective function. 
This enables the method to reason about the global utility of a path and refine trajectories to maximize their value.
The versatility of the algorithm is demonstrated in two different applications, maximizing a volumetric exploration gain and a 3D reconstruction gain.
We introduced an efficiency-inspired value formulation and 3D reconstruction gain that outperform state of the art methods and do not require additional tuning.
Experiments on a real MAV show the robustness and real-time capabilities of our method.
We make our framework available for adaption to other applications and future research.





{\small
\bibliographystyle{IEEEtran}
\bibliography{IEEEfull,references}

\begin{thebibliography}{10}
\providecommand{\url}[1]{#1}
\csname url@rmstyle\endcsname
\providecommand{\newblock}{\relax}
\providecommand{\bibinfo}[2]{#2}
\providecommand\BIBentrySTDinterwordspacing{\spaceskip=0pt\relax}
\providecommand\BIBentryALTinterwordstretchfactor{4}
\providecommand\BIBentryALTinterwordspacing{\spaceskip=\fontdimen2\font plus
\BIBentryALTinterwordstretchfactor\fontdimen3\font minus
  \fontdimen4\font\relax}
\providecommand\BIBforeignlanguage[2]{{%
\expandafter\ifx\csname l@#1\endcsname\relax
\typeout{** WARNING: IEEEtran.bst: No hyphenation pattern has been}%
\typeout{** loaded for the language `#1'. Using the pattern for}%
\typeout{** the default language instead.}%
\else
\language=\csname l@#1\endcsname
\fi
#2}}

\bibitem{nbvp_bircher}
A.~{Bircher}, M.~{Kamel}, K.~{Alexis}, H.~{Oleynikova}, and R.~{Siegwart},
  ``Receding horizon "next-best-view" planner for 3d exploration,'' in
  \emph{IEEE Int.~Conf.~on Robotics \& Automation}, 2016.

\bibitem{Bircher2018}
A.~Bircher, M.~Kamel, K.~Alexis, H.~Oleynikova, and R.~Siegwart, ``Receding
  horizon path planning for 3d exploration and surface inspection,''
  \emph{Autonomous Robots}, vol.~42, no.~2, pp. 291--306, 2018.

\bibitem{nbvp_object_search}
T.~Dang, C.~Papachristos, and K.~Alexis, ``Autonomous exploration and
  simultaneous object search using aerial robots,'' in \emph{IEEE Aerospace
  Conference}, 2018.

\bibitem{popovic2017ipp}
M.~Popovi{\'c}, G.~Hitz, J.~Nieto, I.~Sa, R.~Siegwart, and E.~Galceran,
  ``Online informative path planning for active classification using uavs,'' in
  \emph{IEEE Int.~Conf.~on Robotics \& Automation}, 2017.

\bibitem{Yoder2016}
L.~Yoder and S.~Scherer, ``Autonomous exploration for infrastructure modeling
  with a micro aerial vehicle,'' in \emph{Field and Service Robotics}, 2016.

\bibitem{song_3d_rec}
S.~{Song} and S.~{Jo}, ``Surface-based exploration for autonomous 3d
  modeling,'' in \emph{IEEE Int.~Conf.~on Robotics \& Automation}, 2018.

\bibitem{history_nbvp}
C.~{Witting}, M.~{Fehr}, R.~{Bähnemann}, H.~{Oleynikova}, and R.~{Siegwart},
  ``History-aware autonomous exploration in confined environments using mavs,''
  in \emph{IEEE/RSJ Int.~Conf.~on Intelligent Robots and Systems}, 2018.

\bibitem{yang2013gaussian}
K.~Yang, S.~Keat~Gan, and S.~Sukkarieh, ``A gaussian process-based rrt planner
  for the exploration of an unknown and cluttered environment with a uav,''
  \emph{Advanced Robotics}, vol.~27, no.~6, pp. 431--443, 2013.

\bibitem{nbvp_uncertainty}
C.~{Papachristos}, S.~{Khattak}, and K.~{Alexis}, ``Uncertainty-aware receding
  horizon exploration and mapping using aerial robots,'' in \emph{IEEE
  Int.~Conf.~on Robotics \& Automation}, 2017.

\bibitem{rrt_star}
S.~Karaman and E.~Frazzoli, ``Sampling-based algorithms for optimal motion
  planning,'' \emph{International Journal of Robotics Research}, vol.~30,
  no.~7, pp. 846--894, 2011.

\bibitem{rt_rrt_star}
K.~Naderi, J.~Rajam\"{a}ki, and P.~H\"{a}m\"{a}l\"{a}inen, ``Rt-rrt*: A
  real-time path planning algorithm based on rrt*,'' in \emph{ACM SIGGRAPH
  Conference on Motion in Games}, 2015.

\bibitem{aut_expl_comparison}
M.~Juli{\'a}, A.~Gil, and O.~Reinoso, ``A comparison of path planning
  strategies for autonomous exploration and mapping of unknown environments,''
  \emph{Autonomous Robot.}, vol.~33, no.~4, pp. 427--444, 2012.

\bibitem{yamauchi1997frontier}
B.~Yamauchi, ``A frontier-based approach for autonomous exploration,''
  \emph{IEEE International Symposium on Computational Intelligence in Robotics
  and Automation}, 1997.

\bibitem{rapid}
T.~{Cieslewski}, E.~{Kaufmann}, and D.~{Scaramuzza}, ``Rapid exploration with
  multi-rotors: A frontier selection method for high speed flight,'' in
  \emph{IEEE/RSJ Int.~Conf.~on Intelligent Robots and Systems}, 2017.

\bibitem{gmm_multirobot_expl}
M.~{Corah}, C.~{O’Meadhra}, K.~{Goel}, and N.~{Michael},
  ``Communication-efficient planning and mapping for multi-robot exploration in
  large environments,'' \emph{IEEE Robotics and Automation Letters}, vol.~4,
  no.~2, pp. 1715--1721, 2019.

\bibitem{charrow_ITplanning}
B.~Charrow, G.~Kahn, S.~Patil, S.~Liu, K.~Goldberg, P.~Abbeel, N.~Michael, and
  V.~Kumar, ``Information-theoretic planning with trajectory optimization for
  dense 3d mapping.'' in \emph{Proc.~of Robotics: Science and Systems (RSS)},
  2015.

\bibitem{meng_2stage_expl}
Z.~{Meng}, H.~{Qin}, Z.~{Chen}, X.~{Chen}, H.~{Sun}, F.~{Lin}, and M.~H.~A.
  {Jr}, ``A two-stage optimized next-view planning framework for 3-d unknown
  environment exploration, and structural reconstruction,'' \emph{IEEE Robotics
  and Automation Letters}, vol.~2, no.~3, pp. 1680--1687, 2017.

\bibitem{Selin_nbv_fron}
M.~{Selin}, M.~{Tiger}, D.~{Duberg}, F.~{Heintz}, and P.~{Jensfelt},
  ``Efficient autonomous exploration planning of large-scale 3-d
  environments,'' \emph{IEEE Robotics and Automation Letters}, vol.~4, no.~2,
  pp. 1699--1706, 2019.

\bibitem{marching_cubes}
W.~E. Lorensen and H.~E. Cline, ``Marching cubes: A high resolution 3d surface
  construction algorithm,'' in \emph{ACM SIGGRAPH Computer Graphics}, 1987.

\bibitem{kinect_noise}
C.~V. {Nguyen}, S.~{Izadi}, and D.~{Lovell}, ``Modeling kinect sensor noise for
  improved 3d reconstruction and tracking,'' in \emph{Int.~Conf.~on 3D Imaging,
  Modeling, Processing, Visualization Transmission}, 2012.

\bibitem{realsense_noise}
L.~Keselman, J.~I. Woodfill, A.~Grunnet-Jepsen, and A.~Bhowmik, ``Intel
  realsense stereoscopic depth cameras,'' 2017.

\bibitem{voxblox}
H.~Oleynikova, Z.~Taylor, M.~Fehr, R.~Siegwart, and J.~Nieto, ``Voxblox:
  Incremental 3d euclidean signed distance fields for on-board mav planning,''
  in \emph{IEEE/RSJ Int.~Conf.~on Intelligent Robots and Systems}, 2017.

\bibitem{richter2016polynomial}
C.~Richter, A.~Bry, and N.~Roy, ``Polynomial trajectory planning for aggressive
  quadrotor flight in dense indoor environments,'' in \emph{Robotics
  Research}.\hskip 1em plus 0.5em minus 0.4em\relax Springer, 2016, pp.
  649--666.

\bibitem{safe_local_exploration}
H.~{Oleynikova}, Z.~{Taylor}, R.~{Siegwart}, and J.~{Nieto}, ``Safe local
  exploration for replanning in cluttered unknown environments for microaerial
  vehicles,'' \emph{IEEE Robotics and Automation Letters}, vol.~3, no.~3, 2018.

\bibitem{kamelmpc2016}
M.~Kamel, T.~Stastny, K.~Alexis, and R.~Siegwart, ``Model predictive control
  for trajectory tracking of unmanned aerial vehicles using robot operating
  system,'' in \emph{Robot Operating System (ROS) The Complete Reference,
  Volume 2}, A.~Koubaa, Ed.\hskip 1em plus 0.5em minus 0.4em\relax Springer,
  2017.

\bibitem{rotors_gazebo}
F.~Furrer, M.~Burri, M.~Achtelik, and R.~Siegwart, \emph{Robot Operating System
  (ROS): The Complete Reference (Volume 1)}.\hskip 1em plus 0.5em minus
  0.4em\relax Cham: Springer International Publishing, 2016, ch. RotorS - A
  Modular Gazebo MAV Simulator Framework, pp. 595--625.

\bibitem{unrealcv}
W.~Qiu, F.~Zhong, Y.~Zhang, S.~Qiao, Z.~Xiao, T.~S. Kim, Y.~Wang, and
  A.~Yuille, ``Unrealcv: Virtual worlds for computer vision,'' \emph{ACM
  Multimedia Open Source Software Competition}, 2017.

\end{thebibliography}
}

\end{document}